\documentclass[conference]{IEEEtran}

\usepackage{amssymb}
\usepackage{graphicx}

\usepackage{balance}  
\usepackage{graphics} 
\usepackage{times}    
\usepackage{url}      
\usepackage{graphicx}
\usepackage{array,multirow}
\usepackage{longtable}
\usepackage{tikz}
\usetikzlibrary{positioning}
\usepackage{booktabs}
\usepackage{todonotes}
\usepackage{algorithm, mathtools}
\usepackage{fancyhdr}

\newcommand{\code}[1]{{\lstinline!#1!}}

\newcommand{\argmax}{\operatorname*{arg\,max}}

\usepackage{url}
\urldef{\mailsa}\path|{email1, email2, email3, etc}@essex.ac.uk|  

\begin{document}
%
\title{Population Seeding Techniques for Rolling Horizon Evolution in General Video Game Playing}

\author{\IEEEauthorblockN{Raluca D. Gaina}
\IEEEauthorblockA{University of Essex\\
Colchester, UK\\
Email: rdgain@essex.ac.uk}
\and
\IEEEauthorblockN{Simon M. Lucas}
\IEEEauthorblockA{University of Essex\\
Colchester, UK\\
Email: sml@essex.ac.uk}
\and
\IEEEauthorblockN{Diego P\'erez-Li\'ebana}
\IEEEauthorblockA{University of Essex\\
Colchester, UK\\
Email: dperez@essex.ac.uk}}

\maketitle

\begin{abstract}
While Monte Carlo Tree Search and closely related methods have dominated General Video Game Playing, recent research has demonstrated the promise of Rolling Horizon Evolutionary Algorithms as an interesting alternative. However, there is little attention paid to population initialization techniques in the setting of general real-time video games. Therefore, this paper proposes the use of population seeding to improve the performance of Rolling Horizon Evolution and presents the results of two methods, One Step Look Ahead and Monte Carlo Tree Search, tested on 20 games of the General Video Game AI corpus with multiple evolution parameter values (population size and individual length). An in-depth analysis is carried out between the results of the seeding methods and the vanilla Rolling Horizon Evolution. In addition, the paper presents a comparison to a Monte Carlo Tree Search algorithm. The results are promising, with seeding able to boost performance significantly over baseline evolution and even match the high level of play obtained by the Monte Carlo Tree Search.  
\end{abstract}

          \thispagestyle{plain}
          \fancypagestyle{plain}{
            \fancyhf{} 
            \fancyfoot[L]{978-1-5090-4601-0/17/\$31.00~\copyright2017~IEEE} 
            \renewcommand{\headrulewidth}{0pt}
            \renewcommand{\footrulewidth}{0pt}
          }


%
\IEEEpeerreviewmaketitle

\section{Introduction}

Recent literature features General Video Game Playing (GVGP) more and more, with various researchers using different Game AI frameworks for benchmarking general AI agents \cite{hillclimbing}\cite{Chu2015}\cite{Park2015}. The authors all seem to be in agreement that, although this is a great challenge, its importance is undeniable, exceeding video games.

GVGP is an area focused on developing Artificial Intelligence agents able to achieve a high performance in any previously unknown environment, therefore striving towards General Artificial Intelligence through video games. Games make an excellent domain for testing AI techniques, due to their varying complexity and wide range of problems presented. Additionally, experiments can easily be run multiple times in a constrained scenario, with minimal costs in case of errors and fast feedback, a stark contrast to other areas such as robotics.

This study is carried out using the General Video Game AI \cite{GVGAI} corpus of games, which provides a large collection of interesting and diverse real-time challenges. The GVGAI competition\footnote{www.gvgai.net} has been running for 3 years now and it has increased its coverage in 2016 through a Two Player Planning Track \cite{GVGAI2P} and a Level Generation Track \cite{GVGAILG}. There are several other types of problems in development, such as a Learning Track (which strips the agents of the Forward Model), Screen-Capture Learning or Rule Generation. 

As AI achieves super-human performance in even the most complex of individual games, the more general approach of GVGAI will become increasingly attractive. In 2016, AI became super-human at Go \cite{Silver2016}, and this is speculation on our part, but we believe StarCraft is likely to be dominated by AI within the next few years.

The experiment described in this paper attempts to improve upon a basic Rolling Horizon Evolutionary Algorithm (RHEA) and obtain better performance when tested on a subset of $20$ games of the GVGAI corpus. The proposed technique is focused on generating a better than random initial population from which to start the evolution process i.e. by \emph{seeding the population}. Two different methods are used to this end, a One Step Look Ahead algorithm and Monte Carlo Tree Search, their performance analyzed on multiple RHEA parameter configurations (varying population sizes and individual lengths).

The rest of this document follows a typical structure. Section \ref{sec:research} gives an overview of literature in this domain. Section \ref{sec:background} covers the basic background information necessary on the framework and algorithms used in this study. Section \ref{sec:experiments} reviews the experimental approach and setup. Section \ref{sec:results} reports the results obtained and offers a detailed analysis. Finally, Section \ref{sec:conclusion} wraps up the paper by drawing conclusions and identifying future work.

\section{Relevant Research}\label{sec:research}



Evolutionary Algorithms (EAs) provide a simple, robust and generally applicable approach for searching a wide variety of spaces, and have been the subject of intensive research for more than five decades. In terms of their application to Game AI, much of the effort has been focused on evolving AI agents, or on evolving game content (such as level design) \cite{Togelius2011PCG} \cite{6605565}, game rules or game parameters. Recently, it was shown that Evolutionary Algorithms could be applied as any-time and real-time decision making algorithms for use in Game AI, adopting a similar simulation-driven approach to Monte Carlo Tree Search, while being simpler to implement and offering competitive performance \cite{Perez2013}.  This was initially done for one-player games, but was extended to 2-player games in the form of Rolling Horizon Co-Evolution \cite{Liu2016}.




Various enhancements of EAs have been considered, including hybridization.
One example is the usage of evolution integrated into the Monte Carlo Tree Search simulation step~\cite{Perez2014} or in the roll-out phase to evolve a better policy~\cite{Lucas2014}. Recently, work has moved to incorporate tree structures or Upper Confidence Bounds (UCB) into the evolution instead for a guided and more informed process \cite{BanditRMHC}.

One thing that all Evolutionary Algorithms have in common, regardless of any additional features or the actual evolutionary techniques used, is the initialization of the population. There have been several attempts at exploring this particular improvement. Kazimipour et al. \cite{Kazimipour2014} review all the various methods present in literature and categorize them according to various factors: randomness, compositionality and generality. They identified several techniques which would work in a general environment; however, they suggest that these methods are computationally expensive, therefore not translating well to real time games, for example, which is the domain this paper focuses on.

In addition, Kim et al. \cite{4219044} analyze the effects of initializing an EA population using an optimal solution determined by a Temporal Difference Learning algorithm in the game Othello. This addition appears to lead to a significant improvement in performance and future work in the area is encouraged.

The issue with initializing the population with pseudo-random numbers is raised by Maaranen et al. \cite{Maaranen20041885}, who instead propose a quasi-random sequence method meant to obtain more evenly distributed points in a multi-individual population, in order to better explore the search space. This technique is applied to a genetic algorithm and it is tested on $52$ global optimization problems. Their results are promising, suggesting a higher level of performance over the traditional initialization method.

When it comes to General Video Game Playing, Monte Carlo Tree Search (MCTS) methods have dominated so far and their variations have been explored in various works, as depicted in a survey by Browne et al. \cite{MCTSsurvey}. In the General Video Game AI competition, Open-Loop MCTS emerged as the most powerful method out of the sample controllers provided, standing at the base of multiple participant algorithms and even the winner of the first edition of the competition (ran in 2014), Adrien Cou\"etoux~\cite{GVGAI}. 


However, the Arcade Learning Environment (ALE)~\cite{bellemare13arcade}, still in use by companies such as Google DeepMind~\cite{mnih-dqn-2015}, was one of the first frameworks to allow testing of general agents on video games, presenting the agents with the game through screen capture and requiring an in-game action at every step. Unlike GVGAI, there are limitations to ALE games definition. While the performance achieved by Mnih et al~\cite{mnih-dqn-2015} using Deep Q-Networks (DQN) applied to the ALE environment was impressive, their main goal was to show what could be achieved just be learning to act given the screen capture of the game and a reward function, a process that involves a lengthy training period.


Our interest in the current paper is in methods which can exploit the Forward Model (FM) of the game to achieve intelligent behaviour instantly. Rolling Horizon Evolutionary Algorithms (RHEA) show great promise in this respect. The Forward Model is a game simulator which can be used to rapidly test the consequences of taking a  series of actions, given the current game state. As mentioned above, Perez et al. \cite{Perez2013} tested RHEA techniques on the Physical Salesman Traveling Problem and their results were competitive with MCTS, encouraging research in the area. 

\section{Background}\label{sec:background}

\subsection{General Video Game AI}

GVGAI aims to provide a framework for benchmarking general Artificial Intelligent agents. It currently offers $140$ games in total, $100$ of which are single player and $40$ two-player, some of which stochastic and all real-time. The study in this paper is focused on the single player framework. The games are played by the agents in black box mode, without any knowledge of the rules (e.g. different scoring systems, conditions for ending the game or types of objects in the game - NPCs, portals, resources), but being able to query the current game state for information on game objects. 

In addition, future possible states may be simulated using a Forward Model (FM), which requires an action the agent would wish to perform and returns the game state resulting from that action. However, it is worth noting that any states returned by the FM may not be an accurate representation of the real game due to stochasticity.

The agents have only $40ms$ to make decisions regarding which action to play in the next game tick, except for the initialization step at the start of a game, where they receive $1s$ thinking time. A legal action must be provided at the end of the allocated budget, which may vary in the games between movement or special actions (such as shooting).

\subsection{Evolutionary Algorithms}

The algorithms used in this study are based on the Rolling Horizon Evolutionary Algorithm (RHEA)~\cite{Perez2013}, which encode individuals as sequences of actions. The term ``Rolling Horizon" refers to the fact that the first action of the plan evolved is executed in one game step, then the plan is reevaluated and adjusted, looking one step further into the future, thus slowly expanding the ``horizon". Each individual in the EA is evaluated in a similar manner: the actions are simulated with the use of a Forward Model (FM) following the sequence; the value associated with the state reached at the end (approximated by a heuristic function) is used as the fitness value of the individual.

RHEA's evolution process consists of several iterations (dictated by a fixed number, or a time or memory budget, for example) beginning with population initialization. Subsequently, mutation, tournaments, crossover and other evolutionary methods are used to change individuals and produce new ones. The offspring are evaluated through the steps described above and assigned a fitness value, which leads to the decision of keeping or discarding it in order to move to the next generation with the best individuals found so far. At the end of the process, the algorithm returns the first gene of the best individual in the final population as the action to be played in the game. The evolution is then repeated in the next game tick in the new game state received.



\subsection{Monte Carlo Tree Search}\label{sec:mcts}

Monte Carlo Tree Search (MCTS) is a search-based technique which consists of four steps, iterated over repeatedly until a pre-defined budget is reached (a number of iterations, memory or time, for example). The action returned at the end of the process is the child of the root node considered the best by a recommendation policy (e.g. the most visited child).

In the first step in the process, MCTS selects a non-terminal and not yet fully expanded leaf of the tree via a tree policy. Secondly, a child of the selected node is added to the tree. Thirdly, it simulates ahead, using the new child as the root of a Monte Carlo simulation, with the help of the FM provided by the system. And finally, a heuristic is used to evaluate the state reached at the end of the simulation step and all of the parents of the selected node, up to the root of the tree, are updated with this value. 

The algorithm used in this paper implements an Open Loop variant of this technique, concretely the sample controller from the GVGAI competition. Open Loop means that only statistics and not the actual game states are stored in the nodes of the tree, the FM being used when traversing the tree to simulate the game states. More details of Monte Carlo Tree Search, together with its variations and applications can be found in~\cite{MCTSsurvey}.



\section{Approach and Experimental Setup}\label{sec:experiments}

The aim of this paper is to explore whether initializing the population of an Evolutionary Algorithm with individuals better than random produces an improvement in performance when applied to General Video Game Playing. 

This hypothesis was tested by using 2 different initialization techniques to design variants of the vanilla RHEA, the baseline algorithm in this study, A-Vanilla. Algorithm B-1SLA-S is a seeding variant which employs a One Step Look Ahead technique to select a better starting point in the search space. Algorithm C-MCTS-S uses Monte Carlo Tree Search to seed the RHEA for better analysis of the search space. A fourth algorithm's performance was compared against the RHEA variants, an Open Loop Monte Carlo Tree Search (algorithm D-MCTS), in the simple implementation of the GVGAI competition sample controller.

The effect of the initialization techniques was tested on different configurations of the RHEA algorithm, with population sizes ($P$) and individual lengths ($L$) in the subsequent ranges: $P = \{1, 2, 5, 10, 15, 20\}$, $L = \{6, 8, 10, 14, 16, 20\}$, following the diagonal of the matrix these values would form. In the case of algorithm D-MCTS, its roll-out depth was kept the same as RHEA individual length in order to make the approaches comparable. The largest value tested was 20 due to the fact that, by allowing half of the budget for MCTS computation in algorithm C-MCTS-S, higher values for population size and individual length would result in the algorithm not being able to evaluate even $1$ whole population in the initialization step. 


\subsection{Games}\label{sec:games}

All of the algorithms were tested on the same subset of $20$ single-player games of the current GVGAI corpus. As the aim was to observe performance in different game types, two classifications were used in order to determine a set of games fit for this experiment. Mark Nelson presented a large scale analysis of a basic Monte Carlo Tree Search algorithm in $62$ games, which were sorted based on the performance of this algorithm~\cite{MCTSScaling:CIG16}. Bontrager et al. used clustering techniques on $49$ GVGAI games~\cite{Bontrager2016} based on various features to obtain rough groups of similar games. The $20$ games selected for this experiment were uniformly sampled from both works for a balanced set of $10$ stochastic and $10$ deterministic games (see Table \ref{tab:games} for indices, names and types of these games, as used in the rest of this paper).

\begin{table}[!t]
\begin{center}
\caption{Names, indexes and types of the $20$ games from the subset selected. Legend: S - Stochastic, D - Deterministic.}
\resizebox{0.49\textwidth}{!}{%
\begin{tabular}{|c|c|c||c|c|c|}
\hline
\textbf{Idx} & \textbf{Name}  & \textbf{Type} & \textbf{Idx} & \textbf{Name}  & \textbf{Type}  \\
\hline\hline
$0$ & Aliens & S & $4$ & Bait & D \\
\hline 
$13$ & Butterflies & S & $15$ & Camel Race & D \\
\hline
$18$ & Chase & D & $22$ & Chopper & S \\ 
\hline
$25$ & Crossfire & S & $29$ & Dig Dug & S \\
\hline
$36$ & Escape & D & $46$ & Hungry Birds & D \\ 
\hline 
$49$ & Infection & S & $50$ & Intersection & S \\
\hline
$58$ & Lemmings & D & $60$ & Missile Command & D \\
\hline 
$61$ & Modality & D & $67$ & Plaque Attack & D \\
\hline
$75$ & Roguelike & S & $77$ & Sea Quest & S \\ 
\hline 
$84$ & Survive Zombies & S & $91$ & Wait for Breakfast & D \\
\hline
\end{tabular}%
}
\label{tab:games}
\end{center}
\end{table}

In order to account for the stochastic aspect of the algorithms used in this study, as well as half of the games included in the set, each algorithm was run $100$ times on each game ($20$ times on each of the $5$ levels available). The budget offered for decision-making in each game tick was $900$ FM calls, which is the average number of FM calls that A-Vanilla achieves in the $40ms$ of computational time in the complete $100$ games in the GVGAI-1P corpus. The choice of using FM calls instead of CPU time was made in order to ensure that variations on the machine used for running the experiments would not impact the results, together with the fact that simulating the game ticks is the most expensive part of each algorithm under test.

\subsection{Vanilla RHEA (Algorithm A-Vanilla)}\label{sec:rmhc}

The algorithm described in this subsection is the baseline used in the study and follows the same technique described in \cite{rhanalysis:evoapps17}. It employs a pseudo-random initialization of the population, each gene in the individuals taking on an integer value returned by an RNG (Random Number Generator). Each value is between $0$ and $N-1$ inclusive, where $N$ is the maximum number of legal actions which can be performed from the current game state, therefore the integers mapping to an in-game action.

The evolutionary process continues slightly differently depending on the size of the population. When there is only $1$ individual considered, a new one is mutated at each generation and the individual with the highest fitness value is carried forward to the next iteration. Uniform crossover is introduced for population sizes of $2$ or more and a tournament with size $2$ is used to select the parents of the resulting offspring in the cases where populations contain $3$ or more individuals. The mutation operator is random, the $1$ gene of the individual selected being changed to a different possible value, chosen uniformly at random.

The fitness function consists of a simple heuristic, returning the current game score of the state reached after advancing the Forward Model through all the actions in the individual (or until the end of the game). If an end-game state was reached and it resulted in a loss or a win for the player, the value returned is instead either a large penalty or a large reward, respectively.

\subsection{One Step Look Ahead Seeding (Algorithm B-1SLA-S)}\label{sec:onestep}

The One Step Look Ahead (1SLA) algorithm is a simple technique which exhaustively searches through the actions available from the current state and associates each a $Q$ value, corresponding to the approximated value of the game state reached after performing each action (the value is defined by the same heuristic employed by RHEA). It then selects for execution the action with the highest $Q$ value.

Algorithm B-1SLA-S uses the same evolutionary process as A-Vanilla described above, but the first individual in the initial population is the solution recommended by the 1SLA technique. $L$ iterations of the algorithm are performed, one for each gene in the individual: an exhaustive search is carried out through all of the actions available from the current state, the game state is advanced using the Forward Model, through the best action found and the process is repeated until either the end of the individual or the end of the game is reached. In the second case, the rest of the individual is padded with randomly selected actions.

If the population size is bigger than $1$, the rest of the individuals are obtained by mutating the first individual obtained from the 1SLA algorithm. This method was thought to reduce random bias (the vanilla algorithm potentially not being able to find the current best action because of the random seeding) and to provide a better starting point for evolution.

\subsection{Monte Carlo Tree Search Seeding (Algorithm C-MCTS-S)}

Algorithm C-MCTS-S splits the budget received and uses half of it to first run a Monte Carlo tree search on the current game state, following the steps described in Section \ref{sec:mcts}. The roll-out depth is set to the same value as the individual length in A-Vanilla and the UCB1 formula (with constant $C$ taking the value~\(\sqrt[]{2}\)) is applied as tree policy (see Equation~\ref{eq:ucb1}).

\begin{equation}\label{eq:ucb1}
a^* = \argmax_{a \in A(s)} \left\{Q(s,a) + C \sqrt{\frac{ \ln N(s) }{ N(s,a) }}\right\}
\end{equation}

The first individual in the initial RHEA population is then seeded using the solution recommended by MCTS. Only the first $K$ relevant nodes are selected, by traversing the tree through the most visited actions (the same method used by algorithm D-MCTS when selecting its final action to play). A node is relevant if it has been visited at least $M = 3$ times. The rest of the individual (if any genes have not received a value) is padded with randomly chosen legal actions.

\section{Results and Discussion}\label{sec:results}

The analysis in this section uses a two-tailed Mann-Whitney non-parametric $U$ test to measure the statistical significance of the results for each game ($p$-value~$=0.05$), applied to two performance indicators: win rate and game score achieved. 

In general, both seeding techniques improve the performance of the vanilla algorithm much more when the population size and individual length are small than when they increase. This is in line with the findings in the study performed by Gaina et. al \cite{rhanalysis:evoapps17}, where Random Search (RS) emerged as the best algorithm in the limited budget offered. Therefore, the more the parameter values increase towards RS, the less the impact of the seeding can be observed.

\begin{figure*}[!t]
\centering
\caption{Win rate of algorithms A-Vanilla, B-1SLA-S and C-MCTS-S with configuration $P=1,L=6$ in all $20$ games.}
\includegraphics[width=\textwidth]{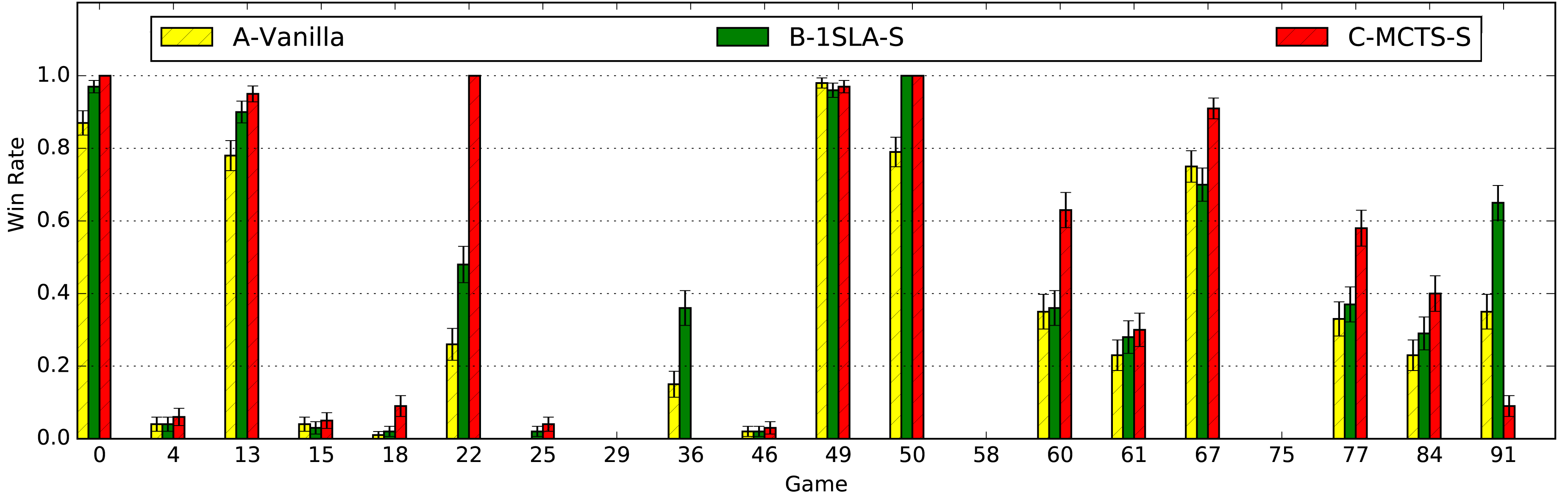}
\label{fig:victories}
\end{figure*}

Table \ref{tab:seeding} presents an overall win rate comparison between the two seeding variants and vanilla RHEA, across all games and configurations. The bottom of the table sums up the number of games in which one algorithm was significantly better than the other two, leading to a total of unique games where a significant improvement was noticed, in all configurations tested. Table \ref{tab:overall1-6} is a complete results example for the configuration $P=1, L=6$ (see Figure \ref{fig:victories} for visualization), the rest of the tables being omitted due to space limitations.

\subsection{Overall Seeding Comparison}

The general trend observed in this study is that the MCTS seeding variant performs significantly better than both algorithms A-Vanilla and B-1SLA-S in $8$ unique games for win rate and $13$ unique games for scores across all configurations, while being significantly worse than either of the other two in only $4$ games for both win rate and score.

It is worth noting that there were a reduced number of games in which A-Vanilla or B-1SLA-S turned out to consistently be significantly better than C-MCTS-S: games with indices 36 and 91 (Escape and Wait for Breakfast, respectively) for both win rate and score and game with index 50 (Intersection) for score only. This is due to the poor performance of MCTS in these games, which is improved in the seeded algorithm over D-MCTS.

In addition, the MCTS seeding shows a steady improvement in several games. The win rates in the games with indices 0, 13 and 22 (Aliens, Butterflies and Chopper, respectively) see an increase to very close to $100\%$ in all configurations. The biggest improvement is observed in game 22, where the A-Vanilla win rate for the smallest configuration ($P=1, L=6$) is only $26\%$ to begin with ($p \ll 0.0001$).

This leads to the conclusion that identifying the type of game being played and applying the correct algorithm seeding and parameters through a meta-heuristic would be highly beneficial to a general AI agent. However, there are also games such as 29 (Dig Dug), 58 (Lemmings) and 77 (Sea Quest) in which the win rate for all algorithms remains at $0\%$, these being particularly difficult games which require greater exploration that neither technique can provide.

\subsection{Pair-wise Seeding Comparison}

\begin{table}[!t]
\centering
\caption{Pair-wise significance comparison between Vanilla RHEA, One Step Look Ahead seeded RHEA and MCTS seeded RHEA.}
\resizebox{0.49\textwidth}{!}{%
\begin{tabular}{|c||c|c|c|c|c|c||c|}
\hline
\textbf{Algorithm} & \textbf{1-6}   & \textbf{2-8} & \textbf{5-10}  & \textbf{10-14} & \textbf{15-16} & \textbf{20-20} & \textbf{Total} \\ \hline\hline
\textbf{A-Vanilla} & 1 (1)            & 0 (0)          & 0 (1)           & \textbf{3 (5)}  &  \textbf{5 (8)}   & \textbf{5 (10)}   & \textbf{8 (11)}  \\ \hline
\textbf{B-1SLA-S} & \textbf{6 (7)} & \textbf{1 (5)}          & \textbf{0 (4)}           & 0 (1)          &  0 (2)            & 0 (2)            & 6 (8)            \\ \hline
\end{tabular}%
}
\smallskip

\resizebox{0.49\textwidth}{!}{%
\begin{tabular}{|c||c|c|c|c|c|c||c|}
\hline
\textbf{A-Vanilla} & 2 (1)            & 2 (3)           & 2 (3)          & \textbf{4 (3)}          & 2 (4) &   \textbf{2 (4)}    & 4 (5)            \\ \hline
\textbf{C-MCTS-S}  & \textbf{10 (16)} & \textbf{6 (11)} & \textbf{4 (7)} & 1 (5) & \textbf{2 (5)}         & 0 (5)  & \textbf{12 (16)} \\ \hline
\end{tabular}%
}
\smallskip

\resizebox{0.49\textwidth}{!}{%
\begin{tabular}{|c||c|c|c|c|c|c||c|}
\hline
\textbf{B-1SLA-S} & 2 (3)           & 2 (4)           & 2 (3)          & 3 (4)          &  2 (4)          &    2 (4)    & 3 (5)            \\ \hline
\textbf{C-MCTS-S}  & \textbf{6 (11)} & \textbf{8 (11)} & \textbf{4 (9)} & \textbf{6 (12)} & \textbf{6 (11)} & \textbf{6 (11)}  & \textbf{10 (13)} \\ \hline
\end{tabular}%
}
\label{tab:comparison}
\end{table}

Pair-wise significance comparison between algorithms A-Vanilla, B-1SLA-S and C-MCTS-S on all the configurations tested can be observed in Table \ref{tab:comparison}. The values represent the number of games (out of $20$ total) in which one algorithm was significantly better than the other regarding victories, as well as scores, in brackets. The totals sum up the unique games in which one algorithm was significantly better than the other across all configurations (maximum of $20$).

\subsubsection{A-Vanilla vs B-1SLA-S}

The One Step Look Ahead seeding appears to produce the best results where the RHEA parameter values are very small (improvements in $6$ games for win rate and $7$ games for score), a change being, however, noticed halfway through the table where the seeding variant actually becomes significantly worse than the vanilla version in up to $5$ games for win rate and $10$ games for score. Overall, across all configurations tested, the 1SLA seeding appears to be worse than the baseline algorithm.

A study of the complete matrix of small parameter values ($P = \{1, 2, 5\}, L=\{6, 8, 10\}$), where the difference in performance is most observed, reveals that the variance in individual length and population size have different effects. On the one hand, increasing the size of the population results in a decrease in the number of games it is significantly better in when compared to A-Vanilla, which is due to the fact that the seeding variant explores the search space much less, with only one optimal solution mutated for all of its individuals. On the other hand, the performance is proportional to the individual length, suggesting that the directed search provided by 1SLA is more effective in cases with big $L$ values compared to A-Vanilla's random sampling. 


\subsubsection{A-Vanilla vs C-MCTS-S}

With parameter values smaller than $P=10, L=14$, the MCTS seeding is significantly better than the vanilla version, the difference being most noticed, again, when the parameter values are small. The decrease in performance is thought to be caused by the roll-out depth of MCTS exceeding the optimal range ($10-12$). Across all configurations, MCTS seeding improves the baseline algorithm in $60\%$ of the games for win rate and $80\%$ for score.

Comparing the complete matrix of small parameter values shows that the population size has a much greater negative effect on the performance than the individual length. The lack of impact of the individual length can be explained by the proportional increase in the roll-out length of MCTS, therefore keeping results comparable. However, the decrease observed with population size increase suggests that the algorithm fails to explore the search space as well as RHEA, therefore balancing of other parameters should be considered.

For configuration $P=5, L=10$, there are two interesting games to look in depth at. In the game with index 77 (Sea Quest), C-MCTS-S increases the win rate of the baseline algorithm from $31\%$ to $68\%$ ($p \ll 0.0001$) and the score from 1225.68 average points to 2485.43 ($p \ll 0.0001$). Another big effect size is perceived in game 15 (Camel Race), in which, although the win rates remain small, there is an increase from $2\%$ to $8\%$ ($p=0.026$). Both games benefit from the balanced exploration and exploitation provided by the MCTS solution which stands at the base of the evolutionary process.

\subsubsection{B-1SLA-S vs C-MCTS-S}

When the two seeding techniques are compared, C-MCTS-S achieves a consistently better performance in $50\%$ of the games for win rate and $65\%$ for scores, whereas being consistently significantly worse in games with indices 36 and 91 (Escape and Wait for Breakfast, respectively) for both win rate and scores and game 50 (Intersection) for score. In game 22 (Chopper), B-1SLA-S with configuration $P=2, L=8$ achieves a $76\%$ win rate, while C-MCTS-S increases it to $100\%$ ($p \ll 0.0001$). In addition, significant improvements with large effect sizes can also be noticed in games 60 (Missile Command, $p=0.016$) and 84 (Survive Zombies, $p=0.033$). However, in the game with index 36 (Escape), in which the MCTS algorithm cannot find a solution, the win rate drops from $30\%$ for B-1SLA-S to $0\%$ for C-MCTS-S ($p \ll 0.0001$).

\subsection{MCTS Comparison}

Although the results indicate algorithm C-MCTS-S to be the best in this setting, an analysis against a pure Monte Carlo Tree Search technique was carried out to validate the findings. This comparison can be seen in Table \ref{tab:olmcts}, in which the values show the number of games in which one algorithm was significantly better than both the others in win rate (and scores, in brackets), adding up to a total of unique games across all configurations. 

The bottom line of Table \ref{tab:olmcts} signifies the amount of games in which, although C-MCTS-S was not the best algorithm, the addition of MCTS seeding to RHEA made it in turn better than the baseline algorithm. This takes into account the cases where C-MCTS-S and D-MCTS were not significantly better than each other, but they still achieved a higher performance than A-Vanilla.

While A-Vanilla consistently obtains significantly more victories and higher scores in its best games (indices 36, 91 and 50), it must be highlighted that the apparent low performance of C-MCTS-S is due to it not being significantly better than D-MCTS. For the direct comparison between C-MCTS-S and A-Vanilla, the reader is referred to Table \ref{tab:comparison}. In this case, the MCTS seeding variant shows improvement over a wider range of games, adding up to $50\%$ games in which a larger win rate was observed and $75\%$ games in which the score increased. The conclusion emerging is that MCTS seeding has a highly beneficial effect, especially in low RHEA parameter values, and further exploration of its advantages is encouraged.

\begin{table}[!t]
\centering
\caption{Significance comparison of algorithms A-Vanilla, C-MCTS-S and D-MCTS in all $20$ games and all configurations.}
\resizebox{.49\textwidth}{!}{%
\begin{tabular}{|c||c|c|c|c|c|c||c|}
\hline
\textbf{Algorithm}                                                   & \textbf{1-6} & \textbf{2-8} & \textbf{5-10} & \textbf{10-14} & \textbf{15-16} & \textbf{20-20} & \textbf{Total} \\ \hline\hline
\textbf{A-Vanilla}                                                   & 2 (1)               & 2 (3)               & 2 (3)                & 2 (3)                 & 2 (4)                 & 2 (4)                 & 3 (4)          \\ \hline
\textbf{C-MCTS-S}                                                    & 0 (2)               & 1 (1)               & 0 (3)                & 0 (0)                 & 0 (1)                 & 0 (1)                 & 1 (7)          \\ \hline
\textbf{D-MCTS}                                                      & 2 (3)               & 0 (0)               & 0 (1)                & 1 (3)                 & 0 (3)                 & 0 (3)                 & 3 (7)          \\ \hline\hline
\textbf{\begin{tabular}[c]{@{}c@{}}Improved \\ Seeding\end{tabular}} & 10 (15)             & 4 (10)              & 2 (4)                & 0 (5)                 & 2 (5)                 & 0 (4)                 & 10 (15)        \\ \hline
\end{tabular}%
}
\label{tab:olmcts}
\end{table}

\begin{table*}[!t]
\begin{center}
\resizebox{\textwidth}{!}{%
\begin{tabular}{|c||c|c|>{\centering\arraybackslash} m{1.7cm}|c|>{\centering\arraybackslash} m{1.7cm}||c|c|>{\centering\arraybackslash} m{1.7cm}|c|>{\centering\arraybackslash} m{1.7cm}|}
\hline
\textbf{Algorithm} & \textbf{Game}  &  \textbf{Victories (\%)} &  \textbf{Significantly better than ...} & \textbf{Scores} &  \textbf{Significantly better than ...} & \textbf{Game}  &  \textbf{Victories (\%)} &  \textbf{Significantly better than ...} & \textbf{Scores} &  \textbf{Significantly better than ...}  \\
\hline\hline
A-Vanilla & \multirow{2}{*}{\textbf{0}} &  $87.00$ $(3.36)$ &  $\O$ & $59.33$ $(1.62)$ &  $\O$ & \multirow{2}{*}{\textbf{49}}& $98.00$ $(1.40)$ &  $\O$ & $11.09$ $(0.61)$ &  $\O$  \\
B-1SLA-S & & $97.00$ $(1.71)$ &  A & $61.95$ $(1.29)$ &  $\O$ & 
& $96.00$ $(1.96)$ &  $\O$ & $11.84$ $(0.71)$ &  $\O$   \\
C-MCTS-S & & $\textbf{100.00 (0.00)}$ &  A, B & $\textbf{68.87 (1.52)}$ &  A, B & 
& $97.00$ $(1.71)$ &  $\O$ & $\textbf{15.25 (0.86)}$ &  A, B  \\
\hline
A-Vanilla &\multirow{2}{*}{\textbf{4}} &  $4.00$ $(1.96)$ &  $\O$ & $2.10$ $(0.29)$ &  $\O$ & 
\multirow{2}{*}{\textbf{50}} & $79.00$ $(4.07)$ &  $\O$ & $-3.03$ $(1.16)$ &  $\O$   \\
B-1SLA-S & & $4.00$ $(1.96)$ &  $\O$ & $3.28$ $(0.51)$ &  $\O$  & 
 & $100.00$ $(0.00)$ &  A & $\textbf{4.25 (0.69)}$ &  A, C  \\
C-MCTS-S &&  $6.00$ $(2.37)$ &  $\O$ & $3.41$ $(0.41)$ &  A  & 
& $100.00$ $(0.00)$ &  A & $1.00$ $(0.00)$ &  A  \\
\hline
A-Vanilla & \multirow{2}{*}{\textbf{13}} &  $78.00$ $(4.14)$ &  $\O$ & $33.12$ $(1.60)$ &  $\O$  & \multirow{2}{*}{\textbf{58}} & $0.00$ $(0.00)$ &  $\O$ & $-9.11$ $(0.37)$ &  $\O$  \\
B-1SLA-S &&  $90.00$ $(3.00)$ &  A & $32.48$ $(1.65)$ &  $\O$  & 
 & $0.00$ $(0.00)$ &  $\O$ & $-0.11$ $(0.05)$ &  A  \\
C-MCTS-S &&  $95.00$ $(2.18)$ &  A & $30.48$ $(1.46)$ &  $\O$  & 
 & $0.00$ $(0.00)$ &  $\O$ & $-0.03$ $(0.02)$ &  A  \\
\hline
A-Vanilla & \multirow{2}{*}{\textbf{15}} &  $4.00$ $(1.96)$ &  $\O$ & $-0.76$ $(0.05)$ &  $\O$  & 
\multirow{2}{*}{\textbf{60}} & $35.00$ $(4.77)$ &  $\O$ & $1.47$ $(0.41)$ &  $\O$  \\
B-1SLA-S &&  $3.00$ $(1.71)$ &  $\O$ & $-0.77$ $(0.05)$ &  $\O$  & 
 & $36.00$ $(4.80)$ &  $\O$ & $1.72$ $(0.43)$ &  $\O$  \\
C-MCTS-S &&  $5.00$ $(2.18)$ &  $\O$ & $-0.75$ $(0.05)$ &  $\O$  & 
& $\textbf{63.00 (4.83)}$ &  A, B & $\textbf{4.65 (0.48)}$ &  A, B  \\
\hline
A-Vanilla & \multirow{2}{*}{\textbf{18}} &  $1.00$ $(0.99)$ &  $\O$ & $2.16$ $(0.20)$ &  $\O$  & 
\multirow{2}{*}{\textbf{61}} & $23.00$ $(4.21)$ &  $\O$ & $0.23$ $(0.04)$ &  $\O$  \\
B-1SLA-S &&  $2.00$ $(1.40)$ &  $\O$ & $2.14$ $(0.22)$ &  $\O$  &
 & $28.00$ $(4.49)$ &  $\O$ & $0.28$ $(0.04)$ &  $\O$  \\
C-MCTS-S &&  $\textbf{9.00 (2.86)}$ &  A, B & $\textbf{3.20 (0.24)}$ &  A, B  &
& $30.00$ $(4.58)$ &  $\O$ & $0.30$ $(0.05)$ &  $\O$  \\
\hline
A-Vanilla & \multirow{2}{*}{\textbf{22}} &  $26.00$ $(4.39)$ &  $\O$ & $2.39$ $(0.62)$ &  $\O$  & 
\multirow{2}{*}{\textbf{67}} & $75.00$ $(4.33)$ &  $\O$ & $35.37$ $(1.60)$ &  $\O$  \\
B-1SLA-S &&  $48.00$ $(5.00)$ &  A & $4.63$ $(0.78)$ &  A  & 
 & $70.00$ $(4.58)$ &  $\O$ & $33.05$ $(1.75)$ &  $\O$  \\
C-MCTS-S &&  $\textbf{100.00 (0.00)}$ &  A, B & $\textbf{16.99 (0.28)}$ &  A, B  & 
& $\textbf{91.00 (2.86)}$ &  A, B & $\textbf{47.17 (1.87)}$ &  A, B  \\
\hline
A-Vanilla & \multirow{2}{*}{\textbf{25}} &  $0.00$ $(0.00)$ &  $\O$ & $-1.01$ $(0.01)$ &  $\O$  & 
\multirow{2}{*}{\textbf{75}} & $0.00$ $(0.00)$ &  $\O$ & $1.60$ $(0.37)$ &  $\O$  \\
B-1SLA-S &&  $2.00$ $(1.40)$ &  $\O$ & $-0.89$ $(0.08)$ &  $\O$  & 
 & $0.00$ $(0.00)$ &  $\O$ & $3.54$ $(0.52)$ &  A  \\
C-MCTS-S &&  $4.00$ $(1.96)$ &  A & $\textbf{0.18 (0.10)}$ &  A, B  & 
& $0.00$ $(0.00)$ &  $\O$ & $5.44$ $(0.62)$ &  A  \\
\hline
A-Vanilla & \multirow{2}{*}{\textbf{29}} & $0.00$ $(0.00)$ &  $\O$ & $5.66$ $(0.76)$ &  $\O$  & 
\multirow{2}{*}{\textbf{77}} & $33.00$ $(4.70)$ &  $\O$ & $903.56$ $(127.82)$ &  $\O$  \\
B-1SLA-S &&  $0.00$ $(0.00)$ &  $\O$ & $9.15$ $(0.77)$ &  A  & 
 & $37.00$ $(4.83)$ &  $\O$ & $1130.36$ $(137.78)$ &  $\O$  \\
C-MCTS-S &&  $0.00$ $(0.00)$ &  $\O$ & $\textbf{14.93 (1.17)}$ &  A, B  & 
& $\textbf{58.00 (4.94)}$ &  A, B & $\textbf{1807.79 (177.44)}$ &  A, B  \\
\hline
A-Vanilla & \multirow{2}{*}{\textbf{36}} & $15.00$ $(3.57)$ &  C & $-0.64$ $(0.07)$ &  $\O$  & 
\multirow{2}{*}{\textbf{84}}& $23.00$ $(4.21)$ &  $\O$ & $0.92$ $(0.38)$ &  $\O$  \\
B-1SLA-S &&  $\textbf{36.00 (4.80)}$ &  A, C & $\textbf{0.34 (0.05)}$ &  A, C  & 
& $29.00$ $(4.54)$ &  $\O$ & $0.92$ $(0.39)$ &  $\O$  \\
C-MCTS-S &&  $0.00$ $(0.00)$ &  $\O$ & $0.00$ $(0.00)$ &  A  & 
& $40.00$ $(4.90)$ &  A & $\textbf{2.27 (0.41)}$ &  A, B  \\
\hline
A-Vanilla & \multirow{2}{*}{\textbf{46}} & $2.00$ $(1.40)$ &  $\O$ & $2.00$ $(1.40)$ &  $\O$  & 
\multirow{2}{*}{\textbf{91}} & $35.00$ $(4.77)$ &  C & $0.35$ $(0.05)$ &  C  \\
B-1SLA-S &&  $2.00$ $(1.40)$ &  $\O$ & $2.00$ $(1.40)$ &  $\O$  & 
& $\textbf{65.00 (4.77)}$ &  A, C & $\textbf{0.65 (0.05)}$ &  A, C  \\
C-MCTS-S &&  $3.00$ $(1.71)$ &  $\O$ & $\textbf{4.60 (1.85)}$ &  A, B  & 
& $9.00$ $(2.86)$ &  $\O$ & $0.09$ $(0.03)$ &  $\O$  \\
\hline
\end{tabular}%
}
\caption{Percentage of victories and average of score achieved (plus standard error) in $20$ different games with configuration $P = 1$ and $L = 6$. Fourth, sixth, ninth and eleventh columns indicate the approaches that are significantly worse than that of the row, using the non-parametric Wilcoxon signed-rank test with p-value $<0.05$.  Bold font for the algorithm that is significantly better than all the other $2$ in either victories or score.}
\label{tab:overall1-6}
\end{center}
\end{table*}

\begin{table*}[!t]
\centering
\resizebox{\textwidth}{!}{%
\begin{tabular}{|c||c|c|c|c|c|c|c||c|}
\hline
\textbf{Game}                   & \textbf{Algorithm} & \textbf{$P=1, L=6$}            & \textbf{$P=2, L=8$}            & \textbf{$P=5, L=10$}              & \textbf{$P=10, L=14$}          & \textbf{$P=15, L=16$}          & \textbf{$P=20, L=20$} & \textbf{Total}   \\ \hline\hline
\multirow{3}{*}{0}     & \textbf{A-Vanilla} & $87.00$ $(3.36)$        & $100.00$ $(0.00)$       & $100.00$ $(0.00)$       & $100.00$ $(0.00)$       & $100.00$ $(0.00)$       & $100.00$ $(0.00)$       & 97.83 (2.16)    \\ \cline{2-9} 
                       & \textbf{B-1SLA-S}  & $97.00$ $(1.71)$        & $100.00$ $(0.00)$       & $100.00$ $(0.00)$       & $100.00$ $(0.00)$       & $100.00$ $(0.00)$       & $100.00$ $(0.00)$       & 99.50 (0.50)    \\ \cline{2-9} 
                       & \textbf{C-MCTS-S}  & \textbf{100.00 (0.00)}  & \textbf{100.00 (0.00)*} & \textbf{100.00 (0.00)*} & \textbf{100.00 (0.00)*} & \textbf{100.00 (0.00)*} & \textbf{100.00 (0.00)*} & 100.00 (0.00)   \\ \hline\hline
\multirow{3}{*}{4}     & \textbf{A-Vanilla} & 4.00 (1.96)             & 5.00 (2.18)             & $9.00$ $(2.86)$         & $8.00$ $(2.71)$         & $11.00$ $(3.13)$        & 5.00 (2.18)             & 7.00 (1.12)     \\ \cline{2-9} 
                       & \textbf{B-1SLA-S}  & 4.00 (1.96)             & 5.00 (2.18)             & $4.00$ $(1.96)$         & $5.00$ $(2.18)$         & $10.00$ $(3.00)$        & 5.00 (2.18)             & 5.5 (1.88)      \\ \cline{2-9} 
                       & \textbf{C-MCTS-S}  & 6.00 (2.37)             & 6.00 (2.37)             & $7.00$ $(2.55)$         & $7.00$ $(2.55)$         & $5.00$ $(2.18)$         & 6.00 (2.37)             & 6.16 (0.33)     \\ \hline\hline
\multirow{3}{*}{13}    & \textbf{A-Vanilla} & 78.00 (4.14)            & 83.00 (3.76)            & $94.00$ $(2.37)$        & $96.00$ $(1.96)$        & $91.00$ $(2.86)$        & $92.00$ $(2.71)$        & 89.00 (2.94)    \\ \cline{2-9} 
                       & \textbf{B-1SLA-S}  & 90.00 (3.00)            & 90.00 (3.00)            & $90.00$ $(3.00)$        & $87.00$ $(3.36)$        & $84.00$ $(3.67)$        & $87.00$ $(3.36)$        & 88.00 (1.04)    \\ \cline{2-9} 
                       & \textbf{C-MCTS-S}  & 95.00 (2.18)            & \textbf{99.00 (0.99)}   & \textbf{100.00 (0.00)}  & $99.00$ $(0.99)$        & \textbf{98.00 (1.40)}   & $97.00$ $(1.71)$        & 98.00 (0.76)    \\ \hline\hline
\multirow{3}{*}{15}    & \textbf{A-Vanilla} & 4.00 (1.96)             & 7.00 (2.55)             & $2.00$ $(1.40)$         & $7.00$ $(2.55)$         & $5.00$ $(2.18)$         & 7.00 (2.55)             & 5.33 (0.77)     \\ \cline{2-9} 
                       & \textbf{B-1SLA-S}  & 3.00 (1.71)             & 3.00 (1.71)             & $3.00$ $(1.71)$         & $7.00$ $(2.55)$         & $7.00$ $(2.55)$         & 2.00 (1.40)             & 4.16 (0.83)     \\ \cline{2-9} 
                       & \textbf{C-MCTS-S}  & 5.00 (2.18)             & 5.00 (2.18)             & $8.00$ $(2.71)$         & $5.00$ $(2.18)$         & $4.00$ $(1.96)$         & $5.00$ $(2.18)$         & 5.33 (0.68)     \\ \hline\hline
\multirow{3}{*}{18}    & \textbf{A-Vanilla} & 1.00 (0.99)             & 2.00 (1.40)             & $7.00$ $(2.55)$         & $6.00$ $(2.37)$         & $8.00$ $(2.71)$         & $6.00$ $(2.37)$         & 5.00 (1.33)     \\ \cline{2-9} 
                       & \textbf{B-1SLA-S}  & 2.00 (1.40)             & 4.00 (1.96)             & $4.00$ $(1.96)$         & $1.00$ $(0.99)$         & $4.00$ $(1.96)$         & 3.00 (1.71)             & 3.00 (0.61)     \\ \cline{2-9} 
                       & \textbf{C-MCTS-S}  & \textbf{9.00 (2.86)}    & \textbf{12.00 (3.25)}   & \textbf{8.00 (2.71)*}   & $11.00$ $(3.13)$        & \textbf{11.00 (3.13)*}  & $11.00$ $(3.13)$        & 10.33 (1.28)    \\ \hline\hline
\multirow{3}{*}{22}    & \textbf{A-Vanilla} & 26.00 (4.39)            & 74.00 (4.39)            & $95.00$ $(2.18)$        & $99.00$ $(0.99)$        & $97.00$ $(1.71)$        & 98.00 (1.40)            & 81.5 (11.74)    \\ \cline{2-9} 
                       & \textbf{B-1SLA-S}  & 48.00 (5.00)            & 76.00 (4.27)            & $72.00$ $(4.49)$        & $33.00$ $(4.70)$        & $24.00$ $(4.27)$        & 13.00 (3.36)            & 44.33 (8.43)    \\ \cline{2-9} 
                       & \textbf{C-MCTS-S}  & \textbf{100.00 (0.00)}  & \textbf{100.00 (0.00)}  & \textbf{100.00 (0.00)}  & \textbf{99.00 (0.99)*}  & \textbf{100.00 (0.00)}  & \textbf{99.00 (0.99)*}  & 99.67 (0.16)    \\ \hline\hline
\multirow{3}{*}{25}    & \textbf{A-Vanilla} & 0.00 (0.00)             & 2.00 (1.40)             & $2.00$ $(1.40)$         & $7.00$ $(2.55)$         & $9.00$ $(2.86)$         & $8.00$ $(2.71)$         & 4.67 (1.39)     \\ \cline{2-9} 
                       & \textbf{B-1SLA-S}  & 2.00 (1.40)             & 3.00 (1.71)             & $4.00$ $(1.96)$         & $4.00$ $(1.96)$         & $3.00$ $(1.71)$         & $6.00$ $(2.37)$         & 3.67 (0.33)     \\ \cline{2-9} 
                       & \textbf{C-MCTS-S}  & \textbf{4.00 (1.96)*}   & \textbf{1.00 (0.99)*}   & \textbf{2.00 (1.40)*}   & \textbf{0.00 (0.00)*}   & \textbf{5.00 (2.18)*}   & \textbf{3.00 (1.71)*}   & 2.50 (0.76)     \\ \hline\hline
\multirow{3}{*}{29}    & \textbf{A-Vanilla} & 0.00 (0.00)             & 0.00 (0.00)             & $0.00$ $(0.00)$         & $0.00$ $(0.00)$         & $0.00$ $(0.00)$         & $0.00$ $(0.00)$         & 0.00 (0.00)     \\ \cline{2-9} 
                       & \textbf{B-1SLA-S}  & 0.00 (0.00)             & 0.00 (0.00)             & $0.00$ $(0.00)$         & $0.00$ $(0.00)$         & $0.00$ $(0.00)$         & $0.00$ $(0.00)$         & 0.00 (0.00)     \\ \cline{2-9} 
                       & \textbf{C-MCTS-S}  & \textbf{0.00 (0.00)*}   & \textbf{0.00 (0.00)*}   & $0.00$ $(0.00)$         & $0.00$ $(0.00)$         & $0.00$ $(0.00)$         & $0.00$ $(0.00)$         & 0.00 (0.00)     \\ \hline\hline
\multirow{3}{*}{36}    & \textbf{A-Vanilla} & 15.00 (3.57)            & 32.00 (4.66)            & $33.00$ $(4.70)$        & \textbf{44.00 (4.96)}   & $42.00$ $(4.94)$        & 39.00 (4.88)            & 34.16 (4.56)    \\ \cline{2-9} 
                       & \textbf{B-1SLA-S}  & \textbf{36.00 (4.80)}   & \textbf{30.00 (4.58)*}  & $33.00$ $(4.70)$        & 31.00 (4.62)            & $33.00$ $(4.70)$        & 37.00 (4.83)            & 33.33 (1.11)    \\ \cline{2-9} 
                       & \textbf{C-MCTS-S}  & 0.00 (0.00)             & 0.00 (0.00)             & $0.00$ $(0.00)$         & $0.00$ $(0.00)$         & $0.00$ $(0.00)$         & 1.00 (0.99)             & 0.16 (0.16)     \\ \hline\hline
\multirow{3}{*}{46}    & \textbf{A-Vanilla} & 2.00 (1.40)             & 4.00 (1.96)             & $3.00$ $(1.71)$         & $1.00$ $(0.99)$         & $3.00$ $(1.71)$         & $3.00$ $(1.71)$         & 2.67 (0.42)     \\ \cline{2-9} 
                       & \textbf{B-1SLA-S}  & 2.00 (1.40)             & 0.00 (0.00)             & $4.00$ $(1.96)$         & $2.00$ $(1.40)$         & $1.00$ $(0.99)$         & 1.00 (0.99)             & 1.67 (0.53)     \\ \cline{2-9} 
                       & \textbf{C-MCTS-S}  & \textbf{3.00 (1.71)*}   & 5.00 (2.18)             & $4.00$ $(1.96)$         & $6.00$ $(2.37)$         & $7.00$ $(2.55)$         & $4.00$ $(1.96)$         & 4.83 (0.57)     \\ \hline\hline
\multirow{3}{*}{49}    & \textbf{A-Vanilla} & 98.00 (1.40)            & 98.00 (1.40)            & $96.00$ $(1.96)$        & $98.00$ $(1.40)$        & $96.00$ $(1.96)$        & $98.00$ $(1.40)$        & 97.33 (0.57)    \\ \cline{2-9} 
                       & \textbf{B-1SLA-S}  & 96.00 (1.96)            & 97.00 (1.71)            & $97.00$ $(1.71)$        & $98.00$ $(1.40)$        & $97.00$ $(1.71)$        & $99.00$ $(0.99)$        & 97.33 (0.33)    \\ \cline{2-9} 
                       & \textbf{C-MCTS-S}  & \textbf{97.00 (1.71)*}  & \textbf{100.00 (0.00)*} & \textbf{94.00 (2.37)*}  & $100.00$ $(0.00)$       & $97.00$ $(1.71)$        & $100.00$ $(0.00)$       & 98.00 (1.00)    \\ \hline\hline
\multirow{3}{*}{50}    & \textbf{A-Vanilla} & 79.00 (4.07)            & 100.00 (0.00)           & $100.00$ $(0.00)$       & $100.00$ $(0.00)$       & $100.00$ $(0.00)$       & $100.00$ $(0.00)$       & 96.50 (3.50)    \\ \cline{2-9} 
                       & \textbf{B-1SLA-S}  & \textbf{100.00 (0.00)*} & \textbf{100.00 (0.00)*} & \textbf{100.00 (0.00)*} & \textbf{100.00 (0.00)*} & \textbf{100.00 (0.00)*} & \textbf{100.00 (0.00)*} & 100.00 (0.00)   \\ \cline{2-9} 
                       & \textbf{C-MCTS-S}  & 100.00 (0.00)           & 100.00 (0.00)           & $100.00$ $(0.00)$       & $100.00$ $(0.00)$       & $100.00$ $(0.00)$       & $100.00$ $(0.00)$       & 100.00 (0.00)   \\ \hline\hline
\multirow{3}{*}{58}    & \textbf{A-Vanilla} & 0.00 (0.00)             & 0.00 (0.00)             & $0.00$ $(0.00)$         & $0.00$ $(0.00)$         & $0.00$ $(0.00)$         & $0.00$ $(0.00)$         & 0.00 (0.00)     \\ \cline{2-9} 
                       & \textbf{B-1SLA-S}  & 0.00 (0.00)             & 0.00 (0.00)             & $0.00$ $(0.00)$         & \textbf{0.00 (0.00)*}   & \textbf{0.00 (0.00)*}   & \textbf{0.00 (0.00)*}   & 0.00 (0.00)     \\ \cline{2-9} 
                       & \textbf{C-MCTS-S}  & 0.00 (0.00)             & 0.00 (0.00)             & \textbf{0.00 (0.00)*}   & $0.00$ $(0.00)$         & $0.00$ $(0.00)$         & $0.00$ $(0.00)$         & 0.00 (0.00)     \\ \hline\hline
\multirow{3}{*}{60}    & \textbf{A-Vanilla} & 35.00 (4.77)            & 46.00 (4.98)            & $51.00$ $(5.00)$        & $57.00$ $(4.95)$        & $58.00$ $(4.94)$        & $59.00$ $(4.92)$        & 51.16 (3.51)    \\ \cline{2-9} 
                       & \textbf{B-1SLA-S}  & 36.00 (4.80)            & 47.00 (4.99)            & $49.00$ $(5.00)$        & $39.00$ $(4.88)$        & $43.00$ $(4.95)$        & $43.00$ $(4.95)$        & 42.83 (2.02)    \\ \cline{2-9} 
                       & \textbf{C-MCTS-S}  & \textbf{63.00 (4.83)}   & \textbf{62.00 (4.85)}   & $59.00$ $(4.92)$        & $67.00$ $(4.70)$        & $69.00$ $(4.62)$        & $64.00$ $(4.80)$        & 64.00 (1.60)    \\ \hline\hline
\multirow{3}{*}{61}    & \textbf{A-Vanilla} & 23.00 (4.21)            & 26.00 (4.39)            & $28.00$ $(4.49)$        & $26.00$ $(4.39)$        & $29.00$ $(4.54)$        & $31.00$ $(4.62)$        & 27.16 (0.87)    \\ \cline{2-9} 
                       & \textbf{B-1SLA-S}  & 28.00 (4.49)            & 22.00 (4.14)            & $23.00$ $(4.21)$        & $25.00$ $(4.33)$        & $21.00$ $(4.07)$        & $32.00$ $(4.66)$        & 25.16 (1.02)    \\ \cline{2-9} 
                       & \textbf{C-MCTS-S}  & 30.00 (4.58)            & 25.00 (4.33)            & $26.00$ $(4.39)$        & $28.00$ $(4.49)$        & $24.00$ $(4.27)$        & $26.00$ $(4.39)$        & 26.50 (0.88)    \\ \hline\hline
\multirow{3}{*}{67}    & \textbf{A-Vanilla} & 75.00 (4.33)            & 79.00 (4.07)            & $89.00$ $(3.13)$        & \textbf{96.00 (1.96)}   & $92.00$ $(2.71)$        & \textbf{96.00 (1.96)*}  & 87.83 (3.24)    \\ \cline{2-9} 
                       & \textbf{B-1SLA-S}  & 70.00 (4.58)            & 72.00 (4.49)            & $85.00$ $(3.57)$        & $89.00$ $(3.13)$        & $83.00$ $(3.76)$        & $82.00$ $(3.84)$        & 80.16 (3.27)    \\ \cline{2-9} 
                       & \textbf{C-MCTS-S}  & \textbf{91.00 (2.86)}   & \textbf{89.00 (3.13)}   & $83.00$ $(3.76)$        & $85.00$ $(3.57)$        & $91.00$ $(2.86)$        & $97.00$ $(1.71)$        & 89.33 (1.32)    \\ \hline\hline
\multirow{3}{*}{75}    & \textbf{A-Vanilla} & 0.00 (0.00)             & 0.00 (0.00)             & $0.00$ $(0.00)$         & $0.00$ $(0.00)$         & $0.00$ $(0.00)$         & $0.00$ $(0.00)$         & 0.00 (0.00)     \\ \cline{2-9} 
                       & \textbf{B-1SLA-S}  & 0.00 (0.00)             & 0.00 (0.00)             & $0.00$ $(0.00)$         & $0.00$ $(0.00)$         & $0.00$ $(0.00)$         & $0.00$ $(0.00)$         & 0.00 (0.00)     \\ \cline{2-9} 
                       & \textbf{C-MCTS-S}  & 0.00 (0.00)             & \textbf{0.00 (0.00)*}   & $0.00$ $(0.00)$         & $0.00$ $(0.00)$         & $0.00$ $(0.00)$         & \textbf{0.00 (0.00)*}   & 0.00 (0.00)     \\ \hline\hline
\multirow{3}{*}{77}    & \textbf{A-Vanilla} & 33.00 (4.70)            & 42.00 (4.94)            & $31.00$ $(4.62)$        & $44.00$ $(4.96)$        & $58.00$ $(4.94)$        & $59.00$ $(4.92)$        & 44.50 (4.16)    \\ \cline{2-9} 
                       & \textbf{B-1SLA-S}  & 37.00 (4.83)            & 44.00 (4.96)            & $48.00$ $(5.00)$        & $34.00$ $(4.74)$        & $52.00$ $(5.00)$        & $43.00$ $(4.95)$        & 43.00 (2.90)    \\ \cline{2-9} 
                       & \textbf{C-MCTS-S}  & \textbf{58.00 (4.94)}   & 51.00 (5.00)            & \textbf{68.00 (4.66)}   & $50.00$ $(5.00)$        & $56.00$ $(4.96)$        & $63.00$ $(4.83)$        & 57.67 (2.69)    \\ \hline\hline
\multirow{3}{*}{84}    & \textbf{A-Vanilla} & 23.00 (4.21)            & 39.00 (4.88)            & $43.00$ $(4.95)$        & $40.00$ $(4.90)$        & $45.00$ $(4.97)$        & $49.00$ $(5.00)$        & 39.83 (4.15)    \\ \cline{2-9} 
                       & \textbf{B-1SLA-S}  & 29.00 (4.54)            & 39.00 (4.88)            & $32.00$ $(4.66)$        & $29.00$ $(4.54)$        & $33.00$ $(4.70)$        & 39.00 (4.88)            & 33.50 (1.68)    \\ \cline{2-9} 
                       & \textbf{C-MCTS-S}  & \textbf{40.00 (4.90)*}  & \textbf{52.00 (5.00)}   & $49.00$ $(5.00)$        & \textbf{41.00 (4.92)*}  & $43.00$ $(4.95)$        & $45.00$ $(4.97)$        & 45.00 (1.92)    \\ \hline\hline
\multirow{3}{*}{91}    & \textbf{A-Vanilla} & 35.00 (4.77)            & 65.00 (4.77)            & $68.00$ $(4.66)$        & $76.00$ $(4.27)$        & $71.00$ $(4.54)$        & $72.00$ $(4.49)$        & 64.50 (6.05)    \\ \cline{2-9} 
                       & \textbf{B-1SLA-S}  & \textbf{65.00 (4.77)}   & 65.00 (4.77)            & $59.00$ $(4.92)$        & $72.00$ $(4.49)$        & $65.00$ $(4.77)$        & $69.00$ $(4.62)$        & 65.83 (2.12)    \\ \cline{2-9} 
                       & \textbf{C-MCTS-S}  & 9.00 (2.86)             & 11.00 (3.13)            & $29.00$ $(4.54)$        & $21.00$ $(4.07)$        & $24.00$ $(4.27)$        & 11.00 (3.13)            & 17.50 (3.13)    \\ \hline\hline
\multirow{3}{*}{Total} & \textbf{A-Vanilla} & 0 (0)                   & 0 (0)                   & 0 (0)                   & \textbf{2 (0)}          & 0 (0)                   & 0 (1)                   & 2 (1)           \\ \cline{2-9} 
                       & \textbf{B-1SLA-S}  & 2 (3)                   & 0 (2)                   & 0 (1)                   & 0 (2)                   & 0 (2)                   & 0 (2)                   & 2 (4)           \\ \cline{2-9} 
                       & \textbf{C-MCTS-S}  & \textbf{6 (11)}         & \textbf{6 (10)}         & \textbf{3 (7)}          & 0 (5)                   & \textbf{2 (4)}          & \textbf{0 (4)}          & \textbf{8 (13)} \\ \hline
\end{tabular}%
}
\caption{Average of victories in all $20$ games. \textbf{Bold style} indicates that the algorithm is significantly better in that game than the other two seeding variants, regarding average victories. \textbf{* symbol} indicates that the algorithm is significantly better in that game than the other two seeding variants, regarding average game score. The bottom of the table adds up the number of games in which the algorithm was significantly better than the other two variants in average victories and average scores in brackets. The non-parametric Wilcoxon signed-rank test with p-value $<0.05$ was used to determine significance.}
\label{tab:seeding}
\end{table*}



\section{Conclusion}\label{sec:conclusion}

This paper presented an experiment focused on observing how a better than random population initialization technique affects the performance of Rolling Horizon Evolutionary Algorithms (RHEA) in General Video Game Playing. Two different seeding techniques were used for testing. First, a One Step Look ahead method, which simply carries out an exhaustive search through all actions available and chooses the best one at each game step. Second, a Monte Carlo Tree Search (MCTS), which took half the budget to process the game from the current state and recommend a solution to serve as a starting point for the evolutionary process. Experiments were carried out in a balanced set of $20$ games of the General Video Game AI framework and using various configurations of RHEA parameters (population size ($P$) and individual length ($L$)).

The results suggest that both seeding variants offer a significant improvement in performance, considering both win rate and in-game score, in particular when the $P$ and $L$ values are small. However, as the parameter values increase, the benefit of seeding decreases, indicating that the unique solution offered by the initialization methods, which the evolution searches around, loses value compared to the wider search space at the disposal of Vanilla RHEA. A conclusion drawn from this is that the seeding directed evolution should be combined with better exploration of the game space in order to achieve optimal results. Nevertheless, as the aim of these algorithms is to attain a high level of play on all games, a positive result on a relatively small sample of games negates the null hypothesis and recommends deeper investigation.

An in-depth comparison between vanilla RHEA, the MCTS seeding algorithm and Open Loop Monte Carlo Tree Search was also performed. The findings of this study pinpoint the fact that, as the evolution parameters increase towards Random Search, so does the performance of RHEA compared to the tree search based methods in several games where the search space is too large for MCTS to traverse efficiently enough. Furthermore, the MCTS seeding does not produce worse results than simply MCTS. Therefore, this seeding technique is shown to have great promise in this environment.

The next steps will be focused on developing the algorithm's exploration of the game space, through further use of tree structures for hybridization, additional roll-outs and circular buffers. Moreover, a wider range of games will be used to ascertain that the difference in performance would indeed be significant in an even more general setting. 

\section*{Acknowledgment}

This work was funded by the EPSRC Centre for Doctoral Training in Intelligent Games  and Game Intelligence (IGGI)  EP/L015846/1.


\bibliographystyle{IEEEtran}
\bibliography{refs}

\vspace{10cm}


\end{document}